\documentclass{article} 
\usepackage{iclr2025_conference,times}
\usepackage[T1]{fontenc}

\usepackage{amsmath,amsfonts,bm}

\def\eqref#1{equation~\ref{#1}}

\def\1{\bm{1}}

\DeclareMathAlphabet{\mathsfit}{\encodingdefault}{\sfdefault}{m}{sl}
\SetMathAlphabet{\mathsfit}{bold}{\encodingdefault}{\sfdefault}{bx}{n}

\DeclareMathOperator*{\argmin}{arg\,min}

\usepackage{hyperref}
\usepackage{url}
\usepackage{booktabs}
\usepackage{graphicx}
\usepackage{cleveref}
\usepackage{multirow}
\usepackage{wrapfig}

\title{\datasetname{}: A Simple But Scalable Reasoning Benchmark}

\author{Jonas Golde\textsuperscript{\dag} \quad Patrick Haller\textsuperscript{\dag} \quad Fabio Barth\textsuperscript{\ddag} \quad Alan Akbik\textsuperscript{\dag}\\ 
\textsuperscript{\dag}Humboldt-Universität zu Berlin \quad \textsuperscript{\ddag}DFKI Berlin\\
\texttt{jonas.max.golde.1@hu-berlin.de}
}

\newcommand{\datasetname}{\textsc{MastermindEval}}

\iclrfinalcopy 
\begin{document}

\maketitle

\begin{abstract}

Recent advancements in large language models (LLMs) have led to remarkable performance across a wide range of language understanding and mathematical tasks. As a result, increasing attention has been given to assessing the true reasoning capabilities of LLMs, driving research into commonsense, numerical, logical, and qualitative reasoning. However, with the rapid progress of reasoning-focused models such as OpenAI's o1 and DeepSeek's R1, there has been a growing demand for reasoning benchmarks that can keep pace with ongoing model developments.
In this paper, we introduce \datasetname{}, a simple, scalable, and interpretable deductive reasoning benchmark inspired by the board game Mastermind. Our benchmark supports two evaluation paradigms: (1) agentic evaluation, in which the model autonomously plays the game, and (2) deductive reasoning evaluation, in which the model is given a pre-played game state with only one possible valid code to infer.
In our experimental results we (1) find that even easy Mastermind instances are difficult for current models and (2) demonstrate that the benchmark is scalable to possibly more advanced models in the future Furthermore, we investigate possible reasons why models cannot deduce the final solution and find that current models are limited in deducing the concealed code as the number of statement to combine information from is increasing.
\end{abstract}

\section{Introduction}

Large language models (LLMs) have demonstrated remarkable performance across various text generation tasks, spanning both text and vision modalities \citep{grattafiori2024llama3herdmodels}. These models, characterized by their large parameter counts, have proven effective in a wide range of language understanding tasks \citep{brown2020languagemodelsfewshotlearners,zhao2024surveylargelanguagemodels}. While LLMs have become increasingly proficient at solving complex reasoning tasks, particularly in mathematical domains, concerns persist regarding their actual ability to perform deductive reasoning. This limitation has drawn significant attention from the research community \citep{sinha-etal-2019-clutrr,geva-etal-2021-aristotle,dziri-et-al-2023-faith}.

\begin{figure}[ht]
\begin{center}
\includegraphics[width=\textwidth]{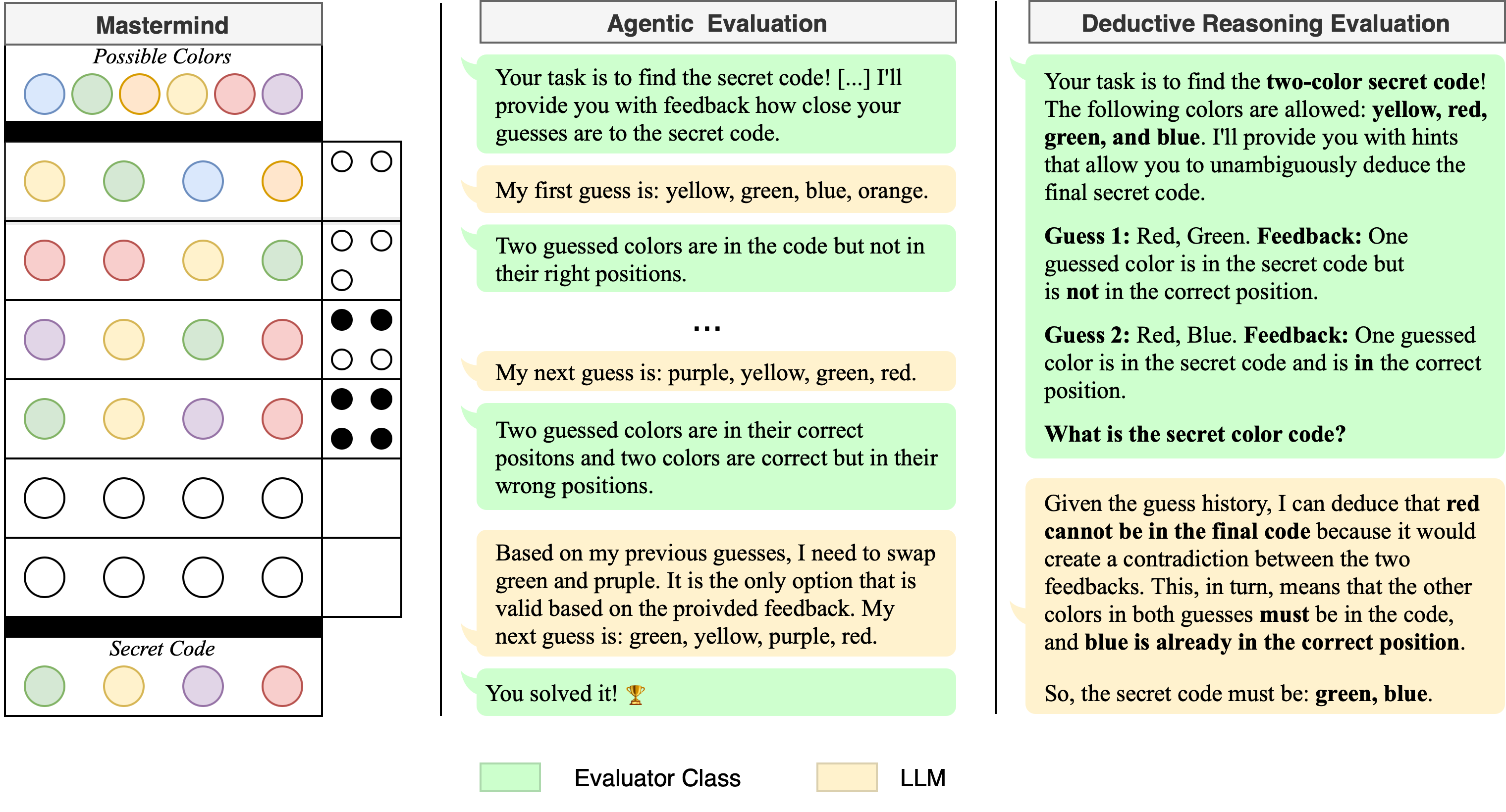}
\end{center}
\caption{\textit{(Left)} A Mastermind game instance with a secret code of length 4 and 6 possible colors, along with corresponding feedback for each guess (black and white pegs). \textit{(Center)} Our strategy evaluation paradigm in \datasetname{}, where the LLM acts as the codebreaker in a multi-turn chat environment. \textit{(Right)} Using Knuth's Five-Guess Algorithm, we pre-play games until only one valid code remains, then prompt the model to deduce the final answer based on the provided hints.}
\label{fig:explanation}
\end{figure}

Several benchmarks have been developed to assess various reasoning capabilities, including LogiQA \citep{liu2020logiqachallengedatasetmachine}, ReClor \citep{yu2020reclorreadingcomprehensiondataset}, and FOLIO \citep{han-etal-2024-folio}. However, many of these high-quality, human-generated benchmarks lack an easy and scalable method for extension. On the other hand, existing logic-based game benchmarks, such as LogicBench \citep{parmar-etal-2024-logicbench}, already incorporate board games to measure reasoning capabilities in language models. Yet, these benchmarks face several limitations.
First, they often involve two-player games, where an LLM's decision-making process is heavily influenced by the opponent’s moves, making it difficult to isolate and interpret the model’s reasoning. Second, these benchmarks are typically not designed for scalability—games like Tic-Tac-Toe do not meaningfully increase in complexity with a larger board, while others, like Mastermind, become disproportionately harder when parameters such as code length are extended. Finally, evaluating models purely based on game outcomes fails to distinguish between genuine strategic reasoning and mere memorization of optimal moves from pretraining data. Furthermore, once these benchmarks are created, they cannot be easily adapted to accommodate future advancements in reasoning models, limiting their long-term applicability.

To address these limitations, we propose \datasetname{}, a benchmark inspired by the game of Mastermind, a logical deduction puzzle in which the player must infer a concealed color code based on feedback from previous guesses. Our benchmark provides two distinct evaluation methodologies: (1) an \textit{agentic evaluation}, where the LLM autonomously plays the game, and (2) a \textit{deductive reasoning evaluation}, where the LLM is tasked to infer the only valid color code in a pre-played game based on prior feedback. We present these variants in~\Cref{fig:explanation}. Additionally, we introduce a multiple-choice derivation of the deductive reasoning evaluation in which the model needs to select the correct answer based on the log-likelihood, thus performing deductive reasoning only using its pretraining objective without relying on newer paradigms such as test-time compute \citep{snell2024scalingllmtesttimecompute,muennighoff2025s1simpletesttimescaling}.

We summarize our contributions as follows:

\begin{enumerate}
    \item We introduce \datasetname{}, a reasoning benchmark based on the game of Mastermind and propose three distinct evaluation methodologies for assessing language model performance using our framework.
    \item We conduct comprehensive experiments on the proposed evaluation methodologies, demonstrating that our benchmark is scalable and highlighting the current limitations of models in combining multiple pieces of information.
    \item We open-source our code and datasets, providing the research community with a scalable reasoning benchmark and potential training datasets to improve reasoning capabilities \footnote{\url{https://github.com/flairNLP/mastermind}}.
\end{enumerate}

\section{Related Work}

Reasoning has long been considered a fundamental capability of language models \citep{ijcai2020p0501,cobbe2021trainingverifierssolvemath,suzgun-etal-2023-challenging,srivastava2023beyond}. As model size increases, reasoning abilities naturally emerge, making it a core attribute of LLMs, as demonstrated in prior work \citep{openai2024gpt4technicalreport}. To better utilize this capability, techniques such as chain-of-thought prompting \citep{wei2023chainofthoughtpromptingelicitsreasoning} and specialized training through reinforcement learning \citep{lambert2025tulu3pushingfrontiers} have been widely adopted. Most recently, many works have demonstrated the positive effect of increased test-time compute (generating more tokens at test time) to be better at reasoning tasks \citep{snell2024scalingllmtesttimecompute,muennighoff2025s1simpletesttimescaling}. Another direction is multi-step reasoning that breaks down complex decision-making and planning tasks into smaller solvable tasks, which are frequently encountered in LLM-based agents \citep{guo2024largelanguagemodelbased,xi2023risepotentiallargelanguage}.

Many benchmarks exist to assess various aspects of reasoning such as inductive reasoning \citep{sinha-etal-2019-clutrr}, temporal reasoning \citep{fatemi2024testtimebenchmarkevaluating,wang-zhao-2024-tram}, spatial reasoning \citep{mirzaee-etal-2021-spartqa}, counterfactual reasoning \citep{tandon-etal-2019-wiqa}, mathematical reasoning \citep{glazer2024frontiermathbenchmarkevaluatingadvanced} or logical reasoning \citep{han-etal-2024-folio}. \citep{gui2024logicgamebenchmarkingrulebasedreasoning}. Additionally, knowledge-based reasoning, particularly commonsense reasoning \citep{talmor2019commonsenseqaquestionansweringchallenge}, assesses whether models can apply commonsense knowledge to make sound inferences. More advanced reasoning tasks, such as theory-of-mind reasoning, evaluate whether models can infer human thoughts and beliefs \citep{wagner2024mindtheorytheorymind}.

Reasoning can also be evaluated through board games such as Chess \citep{CAMPBELL200257} or Go \citep{gopaper}, which require highly complex strategic planning. While specialized models are typically trained for such games, reasoning capabilities in language models are assessed using benchmarks such as LogicGame \citep{gui2024logicgamebenchmarkingrulebasedreasoning}, LogicBench \citep{parmar-etal-2024-logicbench}, BoardgameQA \citep{oh-et-al-2023-boardgameqa}, TruthQuest \citep{mondorf-plank-2024-liar}, and GAMEBoT \citep{lin2024outcomestransparentassessmentllm}. These benchmarks introduce simplified yet challenging board game environments. However, they all share the following limitations:
\textit{(1)} They are not easily scalable in terms of complexity. For example, increasing the board size in Tic-Tac-Toe does not significantly increase its complexity, whereas Mastermind becomes considerably more difficult when using longer secret codes.
\textit{(2)} Evaluating models solely based on game outputs does not determine whether a model is genuinely reasoning and pursuing a strategy to win or merely reproducing good moves memorized from large-scale pretraining data.

\section{MastermindEval} \label{sec:benchmark}

\subsection{Gameplay} \label{sec:gameplay}

In the game of Mastermind, the player (the "codebreaker") attempts to deduce a hidden sequence of symbols by making successive guesses and receiving feedback. The game is defined by three key parameters: the code length $c$, the number of possible symbols $n$, and the allowed number of guesses $g$. A standard game configuration consists of a four-symbol code ($c = 4$) chosen from a set of six possible symbols ($n = 6$), with repetitions allowed, resulting in $6^4 = 1,296$ possible codes.

If the codebreaker's guess is $(x_1, x_2, x_3, x_4)$ and the hidden code is $(y_1, y_2, y_3, y_4)$, the player receives feedback indicating the accuracy of their guess based on two criteria:

\begin{enumerate}
    \item The number of symbols that are in the correct position, i.e., the number of indices $j$ such that $x_j = y_j$ (referred to as "black hits").
    \item The number of symbols that are present in the code but in the wrong position, i.e., the number of indices $j$ where $x_j \neq y_j$ but $x_j = y_k$ for some $k$, provided that $y_k$ has not already been counted as a black hit (these are called "white hits").
\end{enumerate}

As stated in \citet{bonnet2016nash}, this evaluation process can be formalized as a grading function, which is a symmetric function taking two codes as input and returning a pair of integers \( (b, w) \) as output. Given an instance \( (c, n) \) of the Mastermind game, a secret code \( s \), and a guess \( g \), the number of black and white hits is computed as:

$$
b = \sum _{i=1}^{c} \delta (s_i, g_i),
w = \max _{\tilde{g}\in \text{Perm}(g)}\left( \sum _{i=1}^{c} \delta (s_i, \tilde{g}_i)\right) - b,  
$$

where \( \delta \) denotes the Kronecker delta function, which returns 1 if \( s_i = g_i \) and 0 otherwise, and \( Perm(g) \) denotes the set of all permutations of \( g \). 

The objective of the codebreaker is to achieve $c$ black hits by iteratively refining guesses based on the feedback. This iterative process allows the player to eliminate incorrect possibilities and progressively narrow down the valid solutions. In \Cref{fig:explanation} (deductive reasoning view, right side), an example illustrates how feedback systematically reduces the set of possible codes until only one valid solution remains. The challenge is to identify this solution within the allotted number of guesses $g$.

\subsection{Benchmark Construction} \label{sec:benchmark_construction}

We employ an LLM as the codebreaker and implement a dedicated evaluator class that encapsulates the logic of Mastermind, defined by three parameters: code length $c$, number of possible symbols $n$, and the number of allowed guesses $g$. At the start of each game, a secret code is randomly sampled from the $n^c$ possible combinations. The LLM is then prompted to generate an initial guess.

To ensure correct parsing, the model is instructed to clearly indicate its final guess within a structured response format. The guess is extracted using a regular expression to identify and validate the proposed sequence of symbols. It is then evaluated against the secret code, and feedback is provided based on the rules outlined in~\Cref{sec:gameplay} (see also~\Cref{fig:prompt} for reference). Each interaction between the model and the game environment is appended to the overall context, which is provided to the model so that it can consider all previous reasoning when generating subsequent guesses. In the following sections, we introduce the two evaluation paradigms in which LLMs are assessed using \datasetname{}.

\subsection{Agentic Evaluation}
The first evaluation paradigm presents the classical Mastermind game, where the LLM assumes the role of the codebreaker, tasked with deducing a concealed code through iterative (multi-turn) reasoning. The game continues until the model correctly identifies the secret code or reaches a predefined limit on the number of allowed guesses.

Success in this setting depends on refining strategies to maximize information gain with each move. Algorithmic approaches such as those proposed by \citet{Knuth1977TheCA} and \citet{kooi2005yet} effectively balance exploration—formulating diverse guesses to extract informative feedback—and exploitation—leveraging prior feedback to make optimal deductions within known constraints. Strong performance in this paradigm demonstrates an LLM’s ability to perform logical elimination, engage in adaptive reasoning, and efficiently refine hypotheses, all of which are crucial for structured problem-solving tasks.

\subsection{Deductive Reasoning Evaluation}
The agentic evaluation paradigm assesses two distinct capabilities of language models: (1) deducing the final code based on prior guesses and (2) selecting optimal guesses to reach the solution as efficiently as possible. To isolate and evaluate a model's deductive reasoning ability, we introduce a separate evaluation paradigm in which the guessing strategy is predefined.

To achieve this, we construct a dataset comprising over 30,000 pre-played game states across various game configurations, generated using Knuth’s algorithm. Each game progresses until only a single valid code remains based on previous feedback. Unlike the strategy evaluation setting, where models must determine their next move, this paradigm exclusively tests their ability to infer the correct solution given a fixed sequence of prior guesses. The task evaluates whether an LLM can accurately synthesize available information and deduce the unique remaining solution under given constraints.

\noindent\textbf{Knuth's Five-Guess Algorithm.} Knuth’s algorithm \citep{Knuth1977TheCA} solves the Mastermind game optimally in at most five guesses using a minimax strategy. Given a secret code \( s \) from a set \( \mathcal{S} \), the algorithm iteratively selects guesses \( g \in \mathcal{S} \) to minimize the worst-case number of remaining possibilities:

\[
g^* = \argmin_{g \in \mathcal{S}} \max_{\substack{s' \in \mathcal{S} \\ f(g, s') = f(g, s)}} |\mathcal{S}'|
\]

where \( f(g, s) \) denotes the generalized feedback function from~\Cref{sec:gameplay}, and \( \mathcal{S}' \) represents the reduced solution space remaining after incorporating feedback \( f(g, s) \). The process begins with an optimal first guess (e.g., \( (1,1,2,2) \)) and continues until only a single valid code remains (\( |\mathcal{S}| = 1 \)), ensuring a solution within at most five steps.

\subsection{Multiple-Choice Evaluation}
At last, we derive two multiple-choice versions of the deductive reasoning benchmark, inspired by datasets commonly used in the lm-eval-harness \citep{eval-harness} project. In this setting, models rely solely on their pretraining objective to determine the concealed code. We implement two variations: (1) a version where incorrect answers are randomly generated codes, and (2) a version where incorrect answers differ from the correct solution by only one symbol. To ensure fairness, no answer option appears in the previous guesses within the prompt.

\noindent\textbf{Log-Likelihood Prediction.} Given a generated statement, we compute the log-likelihood score \( \log \hat{p}(a_i \mid t) \), where \( t \) represents the prompt and \( a_i \) is an answer candidate. Since the template is the same across all answer options, ranking them is sufficient, allowing us to express the log-likelihood as:

\[
\log \hat{p}(a_i \mid t) = \log \hat{p}(a_i, t) - \log \hat{p}(t).
\]

As causal language models predict token-level log-likelihoods based on prior context, the overall log-likelihood of a sentence is simply the sum of the token-wise log-likelihoods.

\section{Experimental Setup}
We evaluate a diverse set of open-source models on our benchmark, including Qwen-2/2.5 \citep{qwen2025qwen25technicalreport}, Llama-3.1/3.2 \citep{grattafiori2024llama3herdmodels}, Phi-3.5/4 \citep{abdin2024phi3technicalreporthighly,abdin2024phi4technicalreport}, and distilled Qwen-2.5 models from DeepSeek-R1 \citep{deepseekai2025deepseekr1incentivizingreasoningcapability}. In our tables, we denote these models as Qwen-2.5 (R1), covering a parameter range from 1.5B to 7B. All experiments are conducted on four Nvidia H100 GPUs with 80GB of memory. Additionally, we report results for proprietary GPT models, including GPT-4o, GPT-4o-mini, and its reasoning variant, o3-mini \citep{openai2024gpt4technicalreport}.

For the agentic and deductive reasoning evaluations, each model plays 100 games, and we report the solve rate, defined as the fraction of games in which the model correctly identifies the concealed code within the allowed number of guesses. Games where a model correctly guesses the concealed code on the first attempt are excluded from the evaluation. For example, in easy game configurations such as $(c=2, n=4)$, the probability of guessing the concealed code on the first attempt is approximately $6.3\%$. We re-play these games to ensure that every model plays the same set of games. For the deductive reasoning evaluation, we use the same setup, except that the model is limited to a single attempt to solve the game. The only exception is the multiple-choice evaluation, where we assess performance across all 30,000+ instances.

\section{Results}

\subsection{Agentic Evaluation}

\begin{table}
\centering
\begin{tabular}{llcccc}
\toprule
& & \multicolumn{4}{c}{Game Config}\\
Model & Params & ($c=2$, $n=4$) & ($c=3$, $n=5$) & ($c=4$, $n=6$) & ($c=5$, $n=7$) \\
\midrule
Qwen-2 & 1.5B & 0.25 & 0.03 & 0.01 & 0.00 \\
Qwen-2.5 & 3B & 0.69 & 0.12 & 0.02 & 0.00 \\
 & 7B & \underline{0.73} & 0.18 & 0.04 & 0.00 \\
Llama-3.2 & 3B & 0.56 & 0.13 & 0.04 & 0.01 \\
Llama-3.1 & 8B & 0.58 & \underline{0.31} & \underline{0.07} & 0.01 \\
Phi-3.5 (mini) & 3.8B & 0.34 & 0.16 & 0.00 & 0.00 \\
Phi-4 & 14B & \textbf{0.85} & \textbf{0.52} & \textbf{0.30} & \textbf{0.07} \\
Qwen-2.5 (R1) & 1.5B & 0.26 & 0.06 & 0.00 & 0.00 \\
 & 7B & 0.52 & 0.17 & 0.03 & \underline{0.03} \\
\midrule
4o-mini& - & 0.99 & 0.73 & 0.34 & 0.10 \\
4o & - & 1.00 & 0.77 & 0.33 & 0.02 \\
o3-mini & - & 1.00 & 1.00 & 0.86 & 0.92 \\
\bottomrule
\end{tabular}
\caption{\textit{(Agentic Evaluation)} The performance of LLMs playing and solving various game configurations of Mastermind. We present open-source models in the upper part and results for proprietary GPT models in the lower part. We highlight the best scores in bold and the second-best are underlined.}
\label{tab:main_table}
\end{table}

\paragraph{Larger models show better reasoning capabilities.} \Cref{tab:main_table} presents model performance across different game configurations, highlighting trends in capability scaling and efficiency. GPT-4o, GPT-4o-mini, and o3-mini consistently achieve the highest accuracy, performing near-perfectly in simpler settings ($c=2, n=4$) and maintaining strong results even as task complexity increases. Notably, o3-mini, the dedicated reasoning model, achieves high solve rates (over 0.85) even in more challenging settings ($c=5, n=7$). Although not shown in the table, we find that o3-mini's performance declines in more difficult settings, solving 60\% of games when \( c=6, n=8 \) and only 29\% when \( c=7, n=9 \).

Among open-source models, Phi-4 demonstrates the best overall performance, outperforming all other models. However, while it performs well in simpler settings, its accuracy approaches zero as game complexity increases. Furthermore, we observe no significant difference between Qwen-2.5 and its distilled version fine-tuned using DeepSeek-R1, suggesting that the reasoning ability of larger models may not yet be effectively distilled into smaller models.

\begin{wrapfigure}{r}{0.5\textwidth}
    \centering
    \vspace{-10pt}
    \includegraphics[width=0.48\textwidth]{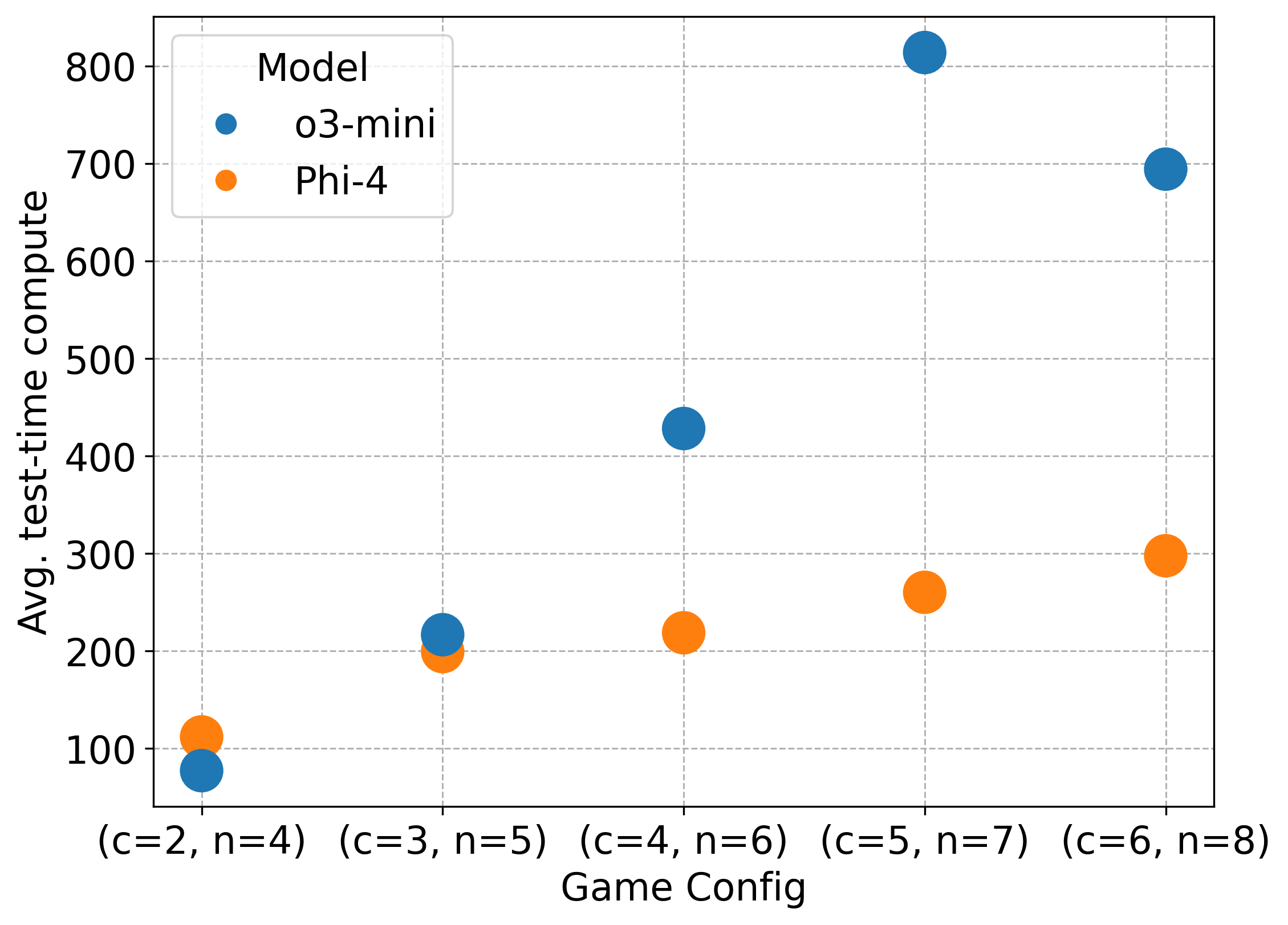}
    \vspace{-10pt} 
    \caption{Test-time compute increases as the model complexity increases indicating \datasetname{} can be easily scaled by increasing the length of the secret code $c$ and the number of possible colors/symbols $n$.}
    \label{fig:testtime}
    \vspace{-10pt} 
\end{wrapfigure}

\paragraph{Test-time compute increases with task complexity.} Additionally, in~\Cref{fig:testtime}, we compare the test-time compute of o3-mini and Phi-4. We observe that both models increase their test-time compute as task complexity grows. However, o3-mini exhibits a much stronger adaptive response, allocating significantly more compute to more difficult tasks. This suggests that o3-mini can reflect on task complexity and dynamically adjust its compute usage to achieve better results, even in the most complex settings (see~\Cref{tab:main_table}).

\paragraph{Models often fail to deduce the secret code, even after eliminating all other possibilities.} We further report scores for ``perfect games'' in~\Cref{tab:perfect_games}, defined as games in which a model has already reduced the remaining state space to a single possible code and then deduces the correct answer in the next guess. We observe that many models struggle to achieve perfect games, often requiring additional guesses even when only one valid code remains. This suggests that models may rely on local optimization rather than a holistic strategy. However, o3-mini consistently achieves high perfect game rates (over 0.78) across various settings, indicating that models explicitly trained for reasoning can learn to think more systematically.

\subsection{Deductive Reasoning Evaluation}

Among open-source models, Phi-4 achieves the highest success rates across all configurations, with scores of 0.50, 0.12, and 0.03, respectively. Llama-3 (8B) and DeepSeek-R1 (7B) also demonstrate competitive performance, ranking among the top models in certain configurations. 

Among proprietary models, o3-mini consistently outperforms all others across every configuration, achieving the highest solve rates. GPT-4o-mini also performs well but lags slightly behind GPT-4o.

Notably, solve rates in this setting are lower compared to the agentic evaluation, indicating that models often fail to deduce the secret code even when doing so is theoretically possible - since the dataset is explicitly designed to ensure solvability. One possible explanation is that models require additional guesses to correctly infer the final code.

\begin{wrapfigure}{r}{0.5\textwidth}
    \centering
    \includegraphics[width=0.48\textwidth]{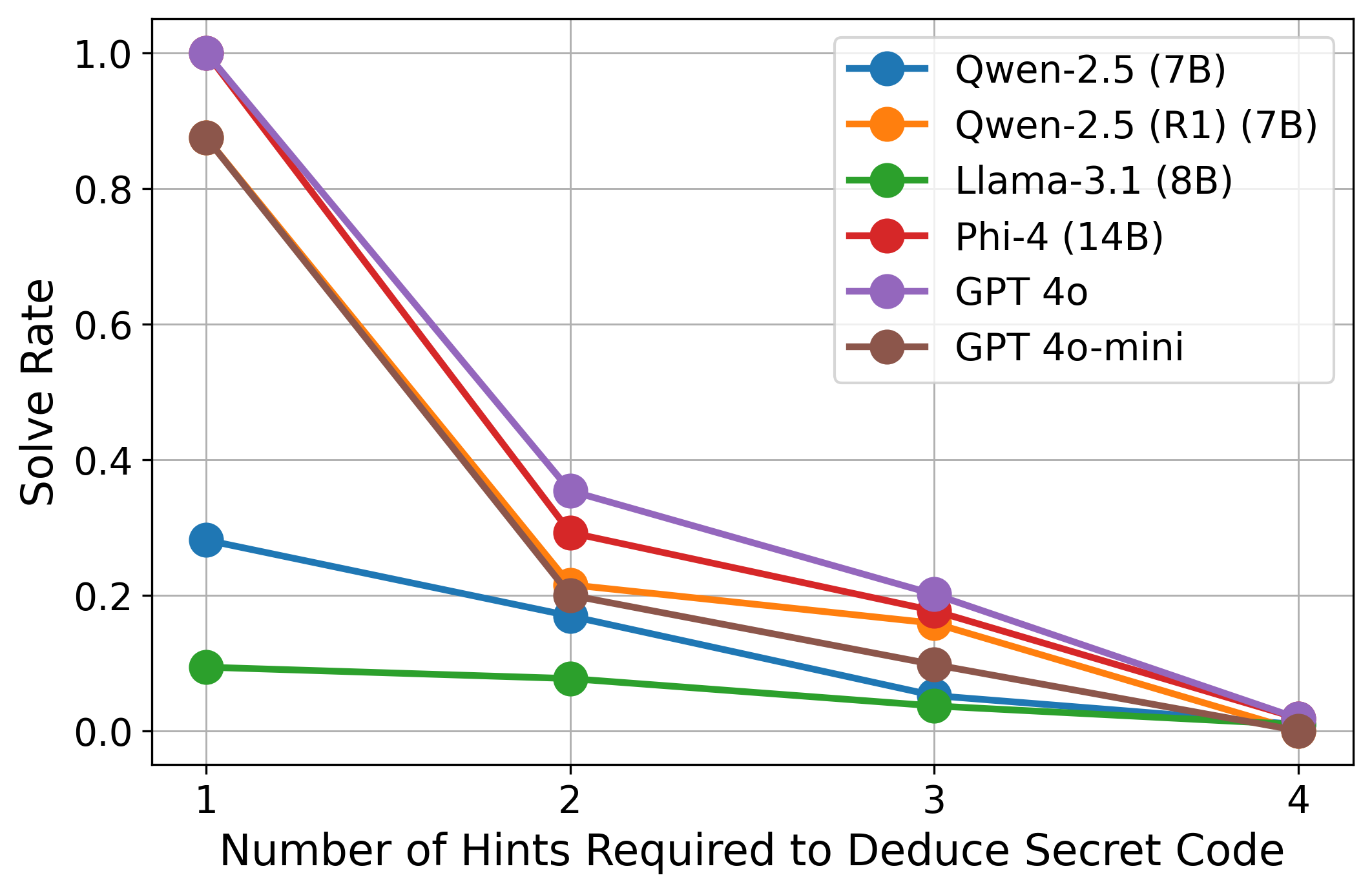}
    \caption{We cluster the solve rate based on the number of hints required to deduce the secret code. We observe that the solve rate approaches 0 when the final code can only be deduced after four chat interactions, even for capable models such as GPT-4o.}
    \label{fig:num_hints}
    \vspace{-20pt}
\end{wrapfigure}

\paragraph{Models struggle when reasoning over a larger amount of information.} \Cref{fig:num_hints} illustrates the number of hints required to deduce the secret code. We observe that models with generally strong performance achieve a high solve rate when only a single hint is provided. However, this rate drops significantly and approaches zero when the model must infer the final code using four guesses. While o3-mini successfully solves all states, we conclude that models explicitly trained for reasoning can effectively integrate information across multiple hints, whereas classical instruction-tuned models struggle in more complex settings.

\begin{table}
\centering
\begin{tabular}{llccc}
\toprule
& & \multicolumn{3}{c}{Game Config}\\
Model & Params & ($c=2$, $n=4$) & ($c=3$, $n=5$) & ($c=4$, $n=6$) \\
\midrule
Qwen-2.5 & 3B & 0.09 & 0.02 & 0.01 \\
 & 7B & 0.25 & 0.04 & \underline{0.02} \\
Llama-3.2 & 3B & 0.08 & 0.04 & \textbf{0.03} \\
Llama-3.1 & 8B & 0.13 & \underline{0.05} & 0.01 \\
Phi-3.5 (mini) & 3.8B & 0.14 & 0.02 & 0.01 \\
Phi-4 & 14B & \textbf{0.50} & \textbf{0.12} & \textbf{0.03} \\
Qwen-2.5 (R1) & 1.5B & 0.07 & 0.00 & 0.00 \\
 & 7B & \underline{0.49} & \underline{0.05} & 0.00 \\
\midrule
GPT-4o-mini & - & 0.34 & 0.09 & 0.00 \\
GPT-4o & - & 0.62 & 0.14 & 0.03 \\
o3-mini & - & 0.99 & 1.00 & 0.99 \\
\bottomrule
\end{tabular}
\caption{\textit{(Deductive Reasoning Evaluation)} The performance of LLMs deducing the secret code in pre-played games. We present results for open-source models in the upper part and proprietary GPT models in the lower part. We indicate best scores in bold and second-best scores underlined.}
\label{tab:deductive_reasoning}
\end{table}

\subsection{Multiple Choice Evaluation}

\Cref{tab:random_answers} and \Cref{tab:close_answers} present the performance of various language models in the multiple-choice evaluation setting, where models rank answer options based on log-likelihood. The two tables correspond to different types of incorrect answers: \Cref{tab:random_answers} features randomly generated codes as distractors, whereas \Cref{tab:close_answers} presents more challenging distractors that differ from the correct code by only a single pin or color.

\paragraph{Performance Across Evaluation Settings.} 
Across both evaluation settings, larger models generally outperform smaller ones, but certain mid-sized models, such as Phi-3.5 (mini) and Qwen-2.5 (7B), achieve competitive results. In particular, Phi-3.5 (mini) achieves the highest scores in most configurations, while Phi-4 performs best in the $(c=3, n=5)$ setting and ranks second in the others. 

A key observation is that model performance is consistently higher when incorrect answers are randomly generated codes rather than subtly modified versions of the correct answer. This suggests that eliminating entirely incorrect codes is relatively easy for language models, whereas distinguishing between highly similar codes is more challenging. This aligns with expectations, as small perturbations in answer choices require models to perform fine-grained probabilistic reasoning rather than relying on obvious likelihood differences.




\begin{table}
\centering
\begin{tabular}{llccc}
\toprule
& & \multicolumn{3}{c}{Game Config} \\
Model & Params & ($c=2$, $n=4$) & ($c=3$, $n=5$) & ($c=4$, $n=6$) \\
\midrule
Qwen-2.5 & 3B & 0.22 & 0.43 & 0.60 \\
 & 7B & 0.31 & 0.48 & 0.70 \\
Llama-3.2 & 3B & 0.23 & 0.40 & 0.56 \\
Llama-3.1 & 8B & 0.30 & 0.42 & 0.60 \\
Phi-3.5 (mini) & 3.8B & \underline{0.38} & \textbf{0.52} & \textbf{0.72} \\
Phi-4 & 14B & \textbf{0.44} & \underline{0.50} & \underline{0.67} \\
Qwen-2.5 (R1) & 1.5B & 0.30 & 0.45 & 0.65 \\
 & 7B & 0.26 & 0.39 & 0.49 \\
\bottomrule
\end{tabular}
\caption{\textit{(Multiple-Choice Evaluation)} Performance of language models in multiple choice evaluation using log-likelihood-based ranking and \textbf{random codes as wrong answer options}. We indicate best scores in bold and second-best scores underlined.}
\label{tab:random_answers}
\end{table}

\begin{table}
\centering
\begin{tabular}{llccc}
\toprule
& & \multicolumn{3}{c}{Game Config} \\
Model & Params & ($c=2$, $n=4$) & ($c=3$, $n=5$) & ($c=4$, $n=6$) \\
\midrule
Qwen-2.5 & 3B & 0.23 & 0.42 & 0.53 \\
 & 7B & 0.37 & 0.48 & 0.57 \\
Llama-3.2 & 3B & 0.26 & 0.42 & 0.53 \\
Llama-3.1 & 8B & 0.35 & 0.45 & 0.56 \\
Phi-3.5 (mini) & 3.8B & \underline{0.38} & \underline{0.50} & \textbf{0.61} \\
Phi-4 & 14B & \textbf{0.49} & \textbf{0.51} & \textbf{0.61} \\
Qwen-2.5 (R1) & 1.5B & 0.35 & 0.43 & 0.58 \\
 & 7B & 0.27 & 0.38 & 0.41 \\
\bottomrule
\end{tabular}
\caption{\textit{(Multiple-Choice Evaluation)} Performance of language models in multiple choice evaluation using log-likelihood-based ranking and \textbf{close codes as wrong answer options}. The wrong answers differ only in one pin compared to the concealed code. We indicate best scores in bold and second-best scores underlined.}
\label{tab:close_answers}
\end{table}

\paragraph{Effect of Game Complexity.} 
Interestingly, we observe an increase in model accuracy as game complexity increases. This trend may seem counterintuitive, as solving more complex instances should generally be harder. However, this phenomenon suggests that in highly constrained settings, language models can leverage statistical regularities from pretraining to make well-informed eliminations. The structured nature of the problem may allow models to reason more effectively using implicit knowledge, even in the absence of explicit step-by-step deductive reasoning.

Nonetheless, it is important to emphasize that this task is inherently different from the agentic evaluation, where models must actively deduce the concealed code by selecting optimal moves. In the multiple-choice setting, the reasoning process is simplified, as models only need to rank pre-given options rather than formulating a search strategy. Thus, while strong performance in this task highlights the effectiveness of pretraining for probabilistic inference, it does not necessarily translate to strategic reasoning capabilities required for interactive gameplay.

\paragraph{Implications for Model Design.} 
These findings suggest that while larger models tend to generalize better, well-optimized mid-sized models can achieve comparable performance in constrained reasoning tasks. Moreover, the challenge of distinguishing between similar answers points to the need for improved fine-grained reasoning mechanisms, possibly through better representation learning or explicit reasoning objectives during training. Future work could explore whether additional reasoning-specific fine-tuning could bridge the performance gap between purely instruction-tuned models and those explicitly trained for structured problem-solving.




\section{Conclusion}

We introduce a benchmark \datasetname{} based on the game of Mastermind to evaluate language models in both strategic gameplay and deductive reasoning. Our results show that model performance declines as game complexity increases, revealing fundamental limitations in multi-step reasoning. Even in simple settings, models struggle to deduce the final code and integrate multiple hints effectively.

Despite these challenges, models perform well in eliminating unlikely answers, likely due to their pretraining objective. However, this ability does not translate into robust logical reasoning in structured problem-solving tasks. These findings suggest that while current models exploit statistical patterns efficiently, significant advancements are needed to develop structured and strategic reasoning capabilities.

\subsubsection*{Acknowledgments}
We thank all reviewers for their valuable comments. Jonas Golde is supported by the Bundesministerium für Bildung und Forschung (BMBF) as part of the project ``FewTuRe'' (project number 01IS24020). Alan Akbik and Patrick Haller are supported by the Deutsche Forschungsgemeinschaft (DFG, German Research Foundation) under Emmy Noether grant ``Eidetic Representations of Natural Language'' (project number 448414230). Further, Alan Akbik is supported under Germany’s Excellence Strategy ``Science of Intelligence'' (EXC 2002/1, project number 390523135). Fabio Barth is supported by the Bundesministerium für Wirtschaft und Energie (BMWi) as part of the project ``OpenGPT-X'' (project number 68GX21007D).

\clearpage

\bibliography{iclr2025_conference}
\bibliographystyle{iclr2025_conference}

\appendix
\section{Appendix}

\subsection{Game Prompt}

\begin{figure}[ht]
\begin{center}
\includegraphics[width=0.9\textwidth]{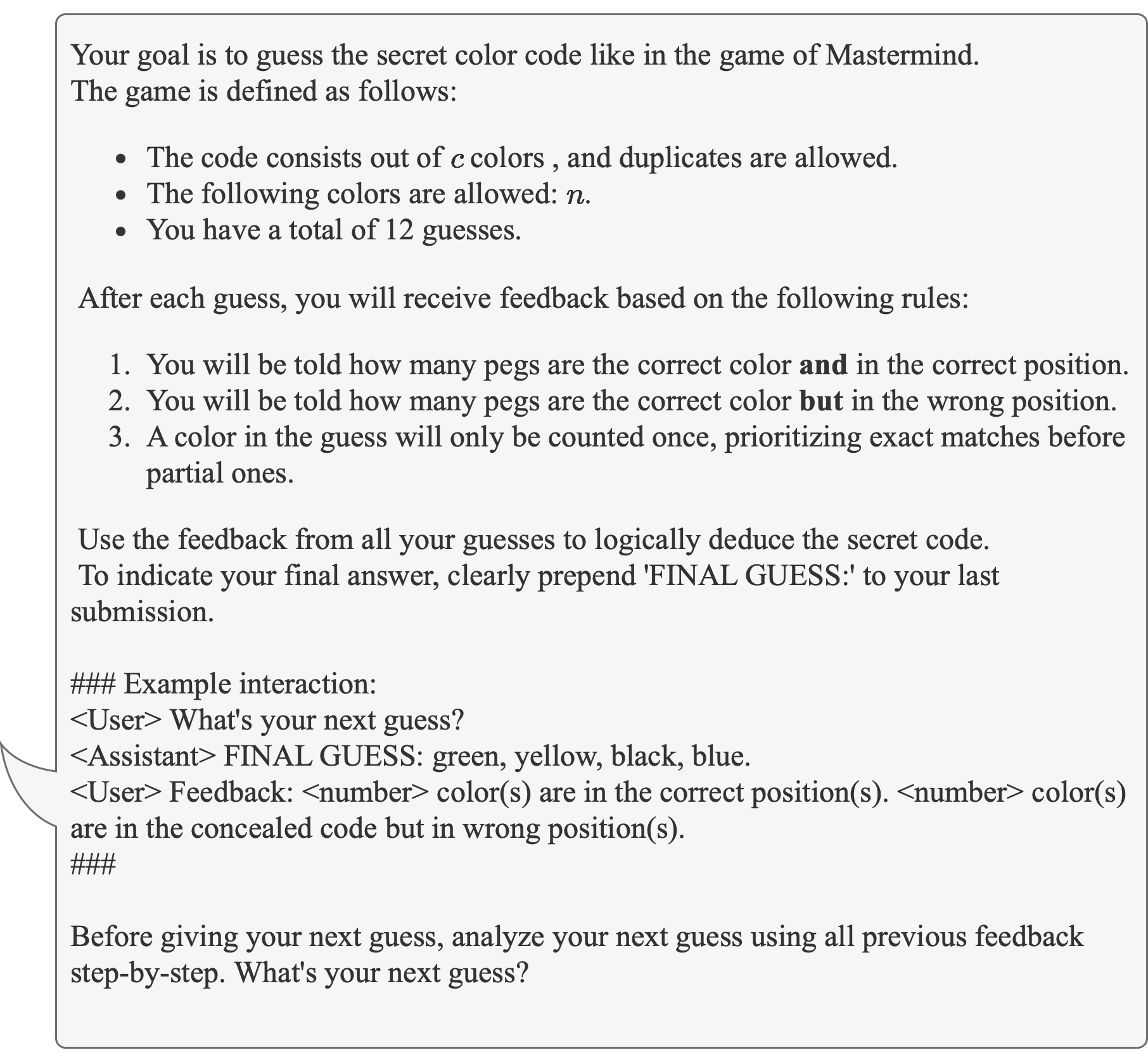}
\end{center}
\caption{The initial prompt used to play a game of Mastermind. We append each answer of the LLM to the prompt such that LLM are enabled to reflect on their own thinking in future steps.}
\label{fig:prompt}
\end{figure}

\subsection{Perfect Games In Agentic Evaluation}

\begin{table}
\centering
\begin{tabular}{llcccc}
\toprule
 & & \multicolumn{4}{c}{Game Config} \\
Model & Params & $(c=2, n=4)$ & $(c=3, n=5)$ & $(c=4, n=6)$ & $(c=5, n=7)$ \\
\midrule
Qwen-2.5 & 1.5B & 0.00 & 0.03 & 0.00 & 0.05 \\
Qwen-2.5 & 3B & 0.06 & 0.01 & 0.03 & 0.09 \\
Qwen-2.5 & 7B & 0.12 & 0.02 & 0.03 & 0.15 \\
DeepSeek-R1 & 1.5B & 0.05 & 0.08 & 0.00 & 0.00 \\
DeepSeek-R1 & 7B & 0.09 & 0.03 & 0.04 & 0.06 \\
DeepSeek-R1 & 14B & 0.26 & 0.00 & - & - \\
Llama-3.1 & 8B & 0.10 & 0.06 & 0.04 & 0.10 \\
Llama-3.2 & 3B & 0.15 & 0.03 & 0.09 & 0.00 \\
Phi-3.5-mini & 3.8B & 0.12 & 0.07 & 0.03 & 0.05 \\
Phi-4 & 14B & 0.42 & 0.21 & 0.10 & 0.09 \\
\midrule
GPT-4o & - & 0.67 & 0.17 & 0.09 & 0.05 \\
GPT-4o-mini & - & 0.33 & 0.12 & 0.08 & 0.02 \\
o3-mini & - & 0.96 & 0.86 & 0.78 & 0.89 \\
\bottomrule
\end{tabular}
\caption{\textit{(Perfect Deduction)} We report the fraction of models that achieve ``perfect games'' in the agentic evaluation paradigm. A perfect game is defined as a sequence of guesses that systematically reduce the state space until only a single possible code remains, followed by a correct deduction on the next guess. If the model requires additional attempts beyond this step, the game is not considered perfect.}
\label{tab:perfect_games}
\end{table}

\section{Limitations and Future Work}
While our benchmark provides valuable insights into the reasoning capabilities of language models through the game of Mastermind, our evaluation has several limitations. Due to computational constraints, we were only able to systematically assess smaller open-source models, restricting our analysis to models that could be run efficiently on available hardware. Additionally, we did not train any models; however, methods such as reinforcement learning via feedback (RLVF) \citep{lambert2025tulu3pushingfrontiers} could potentially improve both strategic play and deductive reasoning in language models.

Another limitation lies in the scope of qualitative information measurement. While we can track the remaining state space in each game configuration using Knuth's algorithm, contributions from other research areas, such as game theory and psychology, could provide a more fine-grained understanding of how models process structured information. In particular, a more precise quantification of the informational content in each hint could further refine our analysis of LLM reasoning capabilities.

Furthermore, we cannot determine whether models are truly reasoning in \datasetname{} or merely relying on pattern-matching learned during pretraining. The paper does not establish whether games of Mastermind are present in the pretraining data, leaving open the possibility that performance is influenced by prior exposure rather than reasoning ability. Verifying this through controlled experiments or data ablation would help clarify the extent to which MastermindEval measures genuine reasoning. Additionally, the study does not analyze how results in MastermindEval correlate with established reasoning benchmarks, making it difficult to assess its validity as a reasoning metric.

Another limitation is the absence of human evaluation. Since performance in MastermindEval appears to correlate with model size, it would be worth investigating whether it also aligns with measures of human intelligence. One straightforward approach would be to test whether people with higher educational attainment perform better in the task. Exploring this connection could provide additional support for MastermindEval as a reasoning benchmark. Future work incorporating human evaluations would help determine whether the task meaningfully reflects reasoning ability beyond model scaling trends.
\end{document}